\documentclass[sigconf]{acmart}

\usepackage{amsmath,amsfonts,bm}

\def\eqref#1{equation~\ref{#1}}

\def\1{\bm{1}}

\def\va{{\bm{a}}}
\def\vb{{\bm{b}}}

\def\ve{{\bm{e}}}

\def\vm{{\bm{m}}}

\def\vp{{\bm{p}}}

\def\vs{{\bm{s}}}

\def\mW{{\bm{W}}}

\DeclareMathAlphabet{\mathsfit}{\encodingdefault}{\sfdefault}{m}{sl}
\SetMathAlphabet{\mathsfit}{bold}{\encodingdefault}{\sfdefault}{bx}{n}

\AtBeginDocument{%
  \providecommand\BibTeX{{%
    \normalfont B\kern-0.5em{\scshape i\kern-0.25em b}\kern-0.8em\TeX}}}

\setcopyright{acmcopyright}
\copyrightyear{2019}
\acmYear{2019}

\acmConference[DLG '19]{The First International Workshop on Deep Learning on Graphs: Methods and Applications}{August 04--08, 2019}{Anchorage, AK}
\acmBooktitle{DLG '19: Deep Learning on Graphs,
  August 04--08, 2019, Anchorage, AK}

\usepackage{subfig}

\begin{document}

\title{Sparse hierarchical representation learning on molecular graphs}


\author{Matthias Bal}
\affiliation{%
  \institution{GTN LTD}
  \city{London}
  \country{UK}
  }
  \authornote{Both authors contributed equally to this research.}
\email{matthias.bal@gtn.ai}

\author{Hagen Triendl}
\affiliation{
  \institution{GTN LTD}
  \city{London}
  \country{UK}}
\authornotemark[1]
\email{hagen.triendl@gtn.ai}

\author{Mariana Assmann}
\affiliation{
  \institution{GTN LTD}
  \city{London}
  \country{UK}}
\email{mariana.assmann@gtn.ai}

\author{Michael Craig}
\affiliation{
  \institution{GTN LTD}
  \city{London}
  \country{UK}}
\email{michael.craig@gtn.ai}

\author{Lawrence Phillips}
\affiliation{
  \institution{GTN LTD}
  \city{London}
  \country{UK}}
\email{lawrence.phillips@gtn.ai}

\author{Jarvist Moore Frost}
\affiliation{
  \institution{GTN LTD}
  \city{London}
  \country{UK}}
\email{jarvist.frost@gtn.ai}

\author{Usman Bashir}
\affiliation{
  \institution{GTN LTD}
  \city{London}
  \country{UK}}
\email{usman.bashir@gtn.ai}

\author{Noor Shaker}
\affiliation{
  \institution{GTN LTD}
  \city{London}
  \country{UK}}
\email{noor@gtn.ai}

\author{Vid Stojevic}
\affiliation{
  \institution{GTN LTD}
  \city{London}
  \country{UK}}
\email{vstojevic@gtn.ai}

\renewcommand{\shortauthors}{Bal and Triendl, et al.}

\begin{abstract}
  Architectures for sparse hierarchical representation learning have recently been proposed for graph-structured data, but so far assume the absence of edge features in the graph. 
We close this gap and propose a method to pool graphs with edge features, inspired by the hierarchical nature of chemistry.
In particular, we introduce two types of pooling layers compatible with an edge-feature graph-convolutional architecture and investigate their performance for molecules relevant to drug discovery on a set of two classification and two regression benchmark datasets of MoleculeNet. 
We find that our models significantly outperform previous benchmarks on three of the datasets and reach state-of-the-art results on the fourth benchmark, with pooling improving performance for three out of four tasks, keeping performance stable on the fourth task, and generally speeding up the training process.
\end{abstract}




\maketitle

\section{Introduction and Related Work}
Predicting chemical properties of molecules has become a prominent application of neural networks in recent years.
A standard approach in chemistry is to conceptualize groups of individual atoms as functional groups with characteristic properties, and infer the properties of a molecule from a multi-level understanding of the interactions between functional groups.
This approach reflects the hierarchical nature of the underlying physics and can be formally understood in terms of renormalization \citep{Lin2017}. 
It thus seems natural to use machine learning models that learn graph representations of chemical space in a local and hierarchical manner. 
This can be realized by coarse-graining the molecular graph in a step-wise fashion, with nodes representing effective objects such as functional groups or rings, connected by effective interactions. 

Much published work leverages node locality by using graph-convolutional networks with message passing to process local information, see \citet{Gilmer2017NeuralMP} for an overview. 
In graph classification and regression tasks, usually, only a global pooling step is applied to aggregate node features into a feature vector for the entire graph \citep{Duvenaud2015ConvolutionalNO, Li2015, Dai2016, Gilmer2017NeuralMP}.\footnote{In some publications \citep{Altae-Tran2017,Li2017} the phrase ``pooling layer'' has been used to refer to a \textsc{max} aggregation step. We reserve the notion of pooling for an operation which creates a true hierarchy of graphs, in line with its usage for images in computer vision.}

An alternative is to aggregate node representations into clusters, which are then represented by a coarser graph \citep{Bruna2013, Niepert2016, Defferrard2016, Monti2016, Simonovsky2017DynamicEF, Fey2017, Mrowca2018}. 
Early work uses predefined and fixed cluster assignments during training, obtained by a clustering algorithm applied to the input graph. 
More recently, dynamic cluster assignments are made on learned node features \citep{Ying2018, Gao2019graph, Cangea2018, Gao2019}. 
A pioneering step in using learnable parameters to cluster and reduce the graph was the \textsc{DiffPool} layer introduced by \citet{Ying2018}. 
Unfortunately, \textsc{DiffPool} is tied to a disadvantageous quadratic memory complexity that limits the size of graphs and cannot be used for large sparse graphs.
A sparse, and therefore more efficient, technique has been proposed by \citet{Gao2019graph} and further tested and explored by \citet{Cangea2018, Gao2019}. 

Sparse pooling layers have so far not been developed for networks on graphs with both node and edge features. This is particularly important for molecular datasets, where edge features may describe different bond types or distances between atoms. When coarsening the molecular graph, new, effective edges need to be created whose edge features represent the effective interactions between the effective nodes.  
In this paper we explore two types of sparse hierarchical representation learning methods for molecules that process edge features differently during pooling: a \emph{simple pooling} layer simply aggregates the features of the involved edges, while a more physically-inspired \emph{coarse-graining} pooling layer determines the effective edge features using neural networks.

We evaluate our approach on established molecular benchmark datasets \citep{Wu2017moleculenet}, in particular on the regression datasets ESOL and lipophilicity and the classification datasets BBBP and HIV, on which various models have been benchmarked \citep{DSNGCN, SMILES2Vec, chemnet, Chemception1, Chemception2, EAGCN, rulebased, Mol2vec, InnerOuterRNNs, potentialnet, SABILSTM, RNNencoder}. We obtain  significantly better results on the datasets ESOL, lipophilicity, and BBBP, and obtain state-of-the-art results on HIV. Simple pooling layers improve results on BBBP and HIV, while coarse-grain pooling improves results on lipophilicity. In general pooling layers can keep performance at least stable while speeding up training.

\section{Approach}
\label{sec:approach}
\subsection{Model architecture}
We represent input graphs in a sparse representation using node ($\va$) and edge ($\ve$) feature vectors
\begin{align}
\va^{(0)}_{i} & = \va_{i}, \enskip i=1, \dots , \; n_{\rm nodes}, \\ \ve^{(0)}_{ij} & = \ve_{ij}, \enskip i,j=1, \dots , n_{\rm nodes} \; \mathrm{for} \; j \in \mathrm{NN}(i),
\end{align}
where $j$ belongs to the set of nearest-neighbours (NN) of $i$. For chemical graphs we encode the atom type as a one-hot vector and its node degree as an additional entry in $\va_{i}$, while the bond type is one-hot encoded in $\ve_{ij}$. Framed in the message-passing framework \citep{Gilmer2017NeuralMP},the graph-convolutional models we use consist of alternating message-passing steps to process information locally and pooling steps that reduce the graph to a simpler sub-graph. Finally, a read-out phase gathers the node features and computes a feature vector for the whole graph that is fed through a simple perceptron layer in the final prediction step.

\noindent{\bf Dual-message graph-convolutional layer} \enskip
Since edge features are an important part of molecular graphs, the model architecture is chosen to give more prominence to edge features.
We design a \emph{dual-message graph-convolutional layer} that supports both node and edge features and treats them similarly. 
First, we compute an aggregate message $\vm^{(k+1)}_{i}$ to a target node from all neighbouring source nodes $j \in {\rm NN}(i)$ using a fully-connected neural network $f_\mW$ acting on the source node features $\va^{(k)}_{i}$ and the edge features $\ve^{(k)}_{ij}$ of the connecting edge. 
A self-message $\vs^{(k+1)}_{i} = \mW_{s}^{(k)}\, \va^{(k)}_{i} + \vb_{s}^{(k)}$ from the original node features is added to the aggregated result. 
New node features are computed by applying batch norm (BN) and a \textsc{ReLU} non-linearity, i.e.\ 
\begin{align}
\vm^{(k+1)}_{i} = \sum_{j \in \mathrm{N}(i)} f_{\mW_a} \left( \va^{(k)}_{j}, \ve^{(k)}_{ij} \right) \, , \\ \tilde \va^{(k+1)}_{i} = \mathrm{ReLU} \left( {\rm BN} \left( \vm^{(k+1)}_{i}  + \vs^{(k+1)}_{i} \right)\right)\, .
\end{align}
In contrast to the pair-message graph-convolutional layer of \citet{Gilmer2017NeuralMP}, we also update the edge feature with the closest node feature vectors via
\begin{align}
\vm^{(k+1)}_{ij} & = g_{\mW_e} \left( \tilde \va^{(k+1)}_{i} + \tilde \va^{(k+1)}_{j}, \ve^{(k)}_{ij} \right), \\ \tilde \ve^{(k+1)}_{ij} & = \mathrm{ReLU} \left( \mathrm{BN} \left( \vm^{(k+1)}_{ij} + \vs^{(k)}_{ij} \right) \right),
\end{align}
where $g$ is a fully-connected neural network and $\vs^{(k)}_{ij} = \mW^{(k)}_{e}\, \ve^{(k)}_{ij} + \vb^{(k)}_{e}$ is the edge feature self-message.

\begin{table*}[ht!]
\small
\centering
\begin{tabular}{lcccc}
 \toprule
  & \multicolumn{2}{c}{\emph{RMSE} results on} & \multicolumn{2}{c}{\emph{ROC-AUC} results on} \\
 \cmidrule(lr){2-3}\cmidrule(lr){4-5}
 \bf{Model} & \emph{ESOL} & \emph{Lipophilicity} & \emph{BBBP} & \emph{HIV} \\
 \midrule
  RF        & 1.07 $\pm$ 0.19      & 0.876 $\pm$ 0.040      & 0.714 $\pm$ 0.000      & --- \\
  Multitask & 1.12 $\pm$ 0.15      & 0.859 $\pm$ 0.013      & 0.688 $\pm$ 0.005      & 0.698 $\pm$ 0.037\\
  XGBoost   & 0.912 $\pm$ 0.000$^a$      & 0.799 $\pm$ 0.054      & 0.696 $\pm$ 0.000      & 0.756 $\pm$ 0.000 \\
  KRR       & 1.53 $\pm$ 0.06      & 0.899 $\pm$ 0.043      & --- & --- \\
  GC        & 0.648 $\pm$ 0.019$^a$       & 0.655 $\pm$ 0.036 & 0.690 $\pm$ 0.009      & 0.763 $\pm$ 0.016\\
  DAG       & 0.82 $\pm$ 0.08      & 0.835 $\pm$ 0.039      & --- & ---\\
  Weave     & 0.553 $\pm$ 0.035$^a$       & 0.715 $\pm$ 0.035      & 0.671 $\pm$ 0.014      & 0.703 $\pm$ 0.039 \\
  MPNN      & 0.58 $\pm$ 0.03 & 0.719 $\pm$ 0.031      & --- & --- \\
  Logreg    &  ---                   & ---                        & 0.699 $\pm$ 0.002      & 0.702 $\pm$ 0.018 \\
  KernelSVM & ---                    & ---                        & \bf{0.729 $\pm$ 0.000} & 0.792 $\pm$ 0.000 \\
  IRV       & ---                    & ---                        & 0.700 $\pm$ 0.000      & 0.737 $\pm$ 0.000 \\
  Bypass    & ---                    & ---                        & 0.702 $\pm$ 0.006      & 0.693 $\pm$ 0.026 \\
  Chemception \citep{Chemception1,Chemception2}   & ---                    & ---                        & ---      & 0.752 \\ 
  Smiles2vec \citep{SMILES2Vec}   &     0.63                & ---                        & ---      & 0.8 \\
  ChemNet \citep{chemnet}   & ---                    & ---                        & ---      & 0.8 \\
  Dummy super node GC \citep{DSNGCN}    & ---                  & ---                        & ---      &  0.766  \\
  EAGCN  \citep{EAGCN}  & ---                    &   \bf{0.61 $\pm$ 0.02}                      & ---      & \bf{0.83 $\pm$ 0.01} \\
  Mol2vec \citep{Mol2vec}   &      0.79               & ---                        & ---      & ---    \\ 
  Outer RNN  \citep{InnerOuterRNNs}  &          0.62           &   0.64                      & ---      & ---   \\
  PotentialNet \citep{potentialnet}   & \bf{0.490 $\pm$ 0.014}              & ---                        & ---      &  ---  \\ 
  SA-BILSTM   \citep{SABILSTM}   & ---             & ---                        & ---      &   \bf{0.83 $\pm$ 0.02}  \\ 
  RNN encoder \citep{RNNencoder}     & 0.58             & 0.62                       &  0.74      &  --- \\ 
  \midrule
  NoPool                   & \bf{0.410 $\pm$ 0.023} & 0.551 $\pm$ 0.010 & 0.846 $\pm$ 0.011 & 0.825 $\pm$ 0.008 \\
 \cmidrule(lr){1-5}
  SimplePooling (0.9)      & \bf{0.410 $\pm$ 0.018} & 0.536 $\pm$ 0.009 & 0.839 $\pm$ 0.022 & 0.824 $\pm$ 0.014 \\
  SimplePooling (0.8)      & 0.417 $\pm$ 0.027 & 0.542 $\pm$ 0.013 & \bf{0.869 $\pm$ 0.010} & 0.816 $\pm$ 0.020 \\
  SimplePooling (0.7)      & 0.485 $\pm$ 0.020 & 0.563 $\pm$ 0.016 & 0.859 $\pm$ 0.009 & 0.825 $\pm$ 0.015 \\
  SimplePooling (0.6)      & 0.413 $\pm$ 0.021 & 0.622 $\pm$ 0.030 & 0.852 $\pm$ 0.006 & \bf{0.840 $\pm$ 0.019} \\
  SimplePooling (0.5)      & 0.437 $\pm$ 0.016 & 0.637 $\pm$ 0.027 & 0.851 $\pm$ 0.012 & 0.822 $\pm$ 0.019 \\
 \cmidrule(lr){1-5}
  CoarseGrainPooling (0.9) & 0.420 $\pm$ 0.015 & \bf{0.517 $\pm$ 0.005} &  0.852 $\pm$ 0.010   & 0.834 $\pm$ 0.015 \\
  CoarseGrainPooling (0.8) & 0.430 $\pm$ 0.019 & 0.529 $\pm$ 0.020 &  0.853 $\pm$ 0.009 & 0.833 $\pm$ 0.009 \\
  CoarseGrainPooling (0.7) & 0.472 $\pm$ 0.013 & 0.530 $\pm$ 0.005 &  0.856 $\pm$ 0.012 & 0.830 $\pm$ 0.007 \\
  CoarseGrainPooling (0.6) & 0.495 $\pm$ 0.053 & 0.536 $\pm$ 0.026 &  0.838 $\pm$ 0.020 & 0.824 $\pm$ 0.026 \\
  CoarseGrainPooling (0.5) & 0.412 $\pm$ 0.031 & 0.535 $\pm$ 0.009 & 0.858 $\pm$ 0.023 & 0.826 $\pm$ 0.010 \\

\bottomrule
\end{tabular}
\caption{(Top) Literature results for the MoleculeNet benchmarks comparing \emph{RMSE} and \emph{ROC-AUC} results, on a range of models. Benchmarks without reference come from \citet{Wu2017moleculenet}, except those values decorated with $^a$, which come from \citet{potentialnet}. 
(Bottom) Our model with coarse-grain pooling, simple pooling, and without pooling. The number in brackets specifies the pooling keep ratio $\rho$ of the pooling layer.
}
\label{molnet-table}
\end{table*}

\noindent{\bf Pooling layer} \enskip
Pooling layers, as introduced in \citet{Gao2019graph}, reduce the number of nodes by a fraction \begin{equation}
\rho = K/n^{(k)}_{\rm nodes},\label{eq:poolfrac}
\end{equation}
specified as a hyperparameter, via scoring all nodes using a learnable projection vector $\vp^{(k)} $, and then selecting the $K$ nodes with the highest score $y^{(k)}_i$. In order to make the projection vector $\vp^{(k)}$ trainable, and thus the node selection differentiable, $\vp^{(k)}$ is also used to determine a gating for each feature vector via
\begin{equation}
y^{(k)}_i=\vp^{(k)} \cdot \tilde \va^{(k)}_{i} 
\, , \qquad \va^{(k)}_{i} = \tilde \va^{(k)}_{i} \tanh\left({y^{(k)}_i}\right)\, ,
\end{equation}
where we only keep the top-$K$ nodes and their gated feature vectors $\va^{(k)}_{i}$.

Pooling nodes requires the creation of new, effective edges between kept nodes while keeping the graph sparse. We discuss in Section~\ref{sec:pool_edgefeats} how to solve this problem in the presence of edge features.

\noindent{\bf Gather layer} \enskip
After graph-convolutional and pooling layers, a graph gathering layer is required to map from node and edge features to a global feature vector. Assuming that the dual-message message-passing steps are powerful enough to distribute the information contained in the edge features to the adjacent node features, we gather over node features only by concatenating \textsc{max} and \textsc{sum}, and acting with a $\mathrm{tanh}$ non-linearity on the result. All models have an additional linear layer that acts on each node individually before applying the gather layer and a final perceptron layer.

\subsection{Pooling with edge features}
\label{sec:pool_edgefeats}
An important step of the pooling process is to create new edges based on the connectivity of the nodes before pooling in order to keep the graph sufficiently connected.
For graphs with edge features this process also has to create new edge features. In addition, the algorithm must be parallel for performance reasons.

\begin{figure}[h]
\begin{center}
\includegraphics[width=0.40\textwidth]{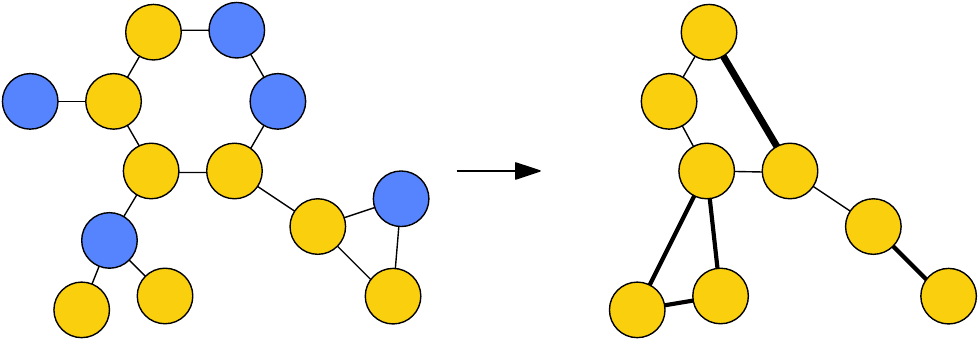}
\end{center}
\caption{Schematic of a graph pooling step (yellow nodes are kept, blue nodes are dropped). Dangling nodes are removed, together with their connecting edges. Pairs of edges connecting a dropped node to two kept nodes are coarse-grained to a new edge (heavy lines). New edges can also be constructed between kept nodes connected by two dropped nodes (heaviest line).\label{fig:pooling}}
\end{figure}

We tackle these issues by specifying how to combine edge features into an effective edge feature between remaining (kept) nodes. If a single dropped node or a pair of dropped nodes connect two kept nodes, we construct a new edge and drop the the ones linked to the dropped nodes. (see Fig.~\ref{fig:pooling}).

We propose two layers to calculate the replacement effective edge feature from the dropped edge features. 
A \emph{simple pooling} layer computes an effective edge-feature by summing all edge feature vectors along the paths connecting pairs of kept nodes. When multiple paths between a pair of nodes are simultaneously reduced, this method will generate overlapping effective edge features. We reduce these to a single vector of the sum of overlapping edge feature vectors. 

We know however that in chemistry effective interactions are more complex functions of the involved component features. Using this as an inspiration, we propose a more expressive \emph{coarse-graining pooling} layer, which is obtained by replacing the simple aggregation function with neural networks to compute effective edge features. In particular, we use two fully-connected neural networks. The first network maps the atom and adjoining edge feature vectors of dropped nodes to a single effective-edge feature. The second network calculates effective edge features for kept edges (between kept nodes) to account for an effective coarse-grained interaction compensating for deleted nodes.

We use pooling layers after every convolutional layer except for the final one. For $N$ convolutional layers, the number of nodes thus gets reduced by a factor $\rho^{N-1}$. This compression not only gets rid of irrelevant information but also reduces memory requirements and makes training faster, as we show in the experiments in Sec.~\ref{sec:exp}.

\section{Experimental Results on MoleculeNet}
\label{sec:exp}

\noindent{\bf Model parameters and implementation} \enskip We use hyperparameter tuning with the hyperband algorithm \citep{JMLR:v18:16-558} to decide on the number of stacks and channel dimensions of graph-convolutional and pooling layers while keeping the pooling keep ratio defined in \eqref{eq:poolfrac} fixed. All our models were implemented in PyTorch and trained on a single Nvidia Tesla K80 GPU using the \textsc{Adam} optimizer with a learning rate of $0.0001$.

\noindent{\bf Evaluation on MoleculeNet} 
\enskip
We evaluate our models with and without pooling layers on the MoleculeNet benchmark set \citep{Wu2017moleculenet}. We focus on four different datasets, comprised of the regression benchmarks ESOL (1128 molecules) and Lipophilicity (4200 molecules), where performance was evaluated by RMSE, and the classification benchmarks on the BBBP (2039 molecules) and HIV (41127 molecules) datasets, evaluated via ROC-AUC. 
Following \citet{Wu2017moleculenet}, we used a scaffold split for the classification datasets as provided by the DeepChem package.
Apart from the benchmarks generated in the original paper, various models have been evaluated on these datasets \citep{DSNGCN, SMILES2Vec, chemnet, Chemception1, Chemception2, EAGCN, rulebased, Mol2vec, InnerOuterRNNs,potentialnet,SABILSTM,RNNencoder}. An overview of the results in the literature can be found in the top of Table~\ref{molnet-table}. 
Our results are the mean and standard deviation of 5 runs over 5 random splits (ESOL, Lipophilicity) or 5 runs over the same scaffold split (BBBP, HIV). Datasets were split into training (80\%), validation (10\%), and held-out test sets (10\%). The validation set was used to tune model hyperparameters. All reported metrics are results on the test set.
The results of our models with and without pooling are displayed in the lower part of the table.

\begin{table}[h]
\begin{tabular}{lccccc}
\toprule
pooling keep ratio & 0.9 & 0.8 & 0.7 & 0.6 & 0.5 \\
\midrule
Speed-up   & 16\% & 24\% & 47\% & 55\% & 70\%\\
\bottomrule
\end{tabular}
\caption{Speed-up of pooling runs of the HIV data set using Simple Pooling. The speed-up is measured as increase in speed in terms of elapsed real time compared to the run without pooling layer.}
\label{speedup}
\end{table}

For the regression tasks, we found that our models significantly outperformed previous models for both datasets, with pooling layers keeping performance stable for ESOL and the coarse-grain pooling layer significantly improving results for Lipophilicity (see Table~\ref{molnet-table}). Regarding classification tasks, we found that our models significantly outperformed previous models on BBBP and also exceeded previous benchmarks for the HIV dataset. For both datasets simple pooling layers improved performance. 
Curiously, the extent to which pooling layers improve performance and which layer is better suited for a particular task strongly depends on the dataset. It seems that simple pooling performs much better for classification tasks while for regression tasks it depends on the dataset.

We also measure the speed-up given by pooling layers during the evaluation on the HIV dataset in terms of elapsed real-time, using the simple pooling layer. The results are displayed in Table~\ref{speedup}. We see significant speed-ups for moderate values of the pooling ratio.

\section{Conclusion}
\label{sec:conclusion}
We introduce two graph-pooling layers for sparse graphs with node and edge features and evaluate their performance on molecular graphs. While our model without pooling significantly outperforms benchmarks on ESOL, lipophilicity and BBBP and reaches state-of-the-art results on HIV in the MoleculeNet dataset, we find that our pooling methods improve performance and provide a speedup of up to 70\% in the training of graph-convolutional neural networks that utilize edge features, along with a reduction in memory requirements.

While all experiments have been performed on datasets comprised of small, druglike molecules, we expect even stronger performance for datasets comprised of larger graphs like protein structures, where pooling can create a large, sequential hierarchy of graphs.
More generally, our work may result in more pertinent and information-effective latent space representations for graph-based machine learning models.

\bibliographystyle{ACM-Reference-Format}
\bibliography{DLG-2019}

\appendix

\section{Supplementary Material}

\subsection{Material science application: Clean Energy Project 2017 dataset}
\label{app:cep}

\begin{table*}
\centering
\begin{tabular}{ccccc}
\toprule

 \bf{pooling}  & \multicolumn{3}{c}{Multi-task} & \multicolumn{1}{c}{Single-task} \\
 \cmidrule(lr){2-4}\cmidrule(lr){5-5}
 \bf{ratio}  & $R^2$ on PCE & $R^2$ on GAP & $R^2$ on HOMO & $R^2$ on PCE \\
 \toprule
 none& 0.862 $\pm$ 0.005 & 0.967 $\pm$ 0.001 & 0.981 $\pm$ 0.000 & 0.866 $\pm$ 0.003 \\
 \midrule
 0.9 & 0.863 $\pm$ 0.003   & 0.966 $\pm$ 0.001 & 0.981 $\pm$ 0.000 & 0.862 $\pm$ 0.002 \\
 0.8 & 0.860 $\pm$ 0.003   & 0.966 $\pm$ 0.001 & 0.981 $\pm$ 0.001 & 0.859 $\pm$ 0.004 \\
 0.7 & 0.856 $\pm$ 0.003   & 0.964 $\pm$ 0.001 & 0.980 $\pm$ 0.001 & 0.855 $\pm$ 0.004 \\
 0.6 & 0.854 $\pm$ 0.007   & 0.962 $\pm$ 0.002 & 0.979 $\pm$ 0.001 & 0.853 $\pm$ 0.003 \\
 0.5 & 0.844 $\pm$ 0.003   & 0.955 $\pm$ 0.002 & 0.974 $\pm$ 0.001 & 0.833 $\pm$ 0.007 \\
 \toprule
             & RMSE on PCE & RMSE on GAP & RMSE on HOMO & RMSE on PCE\\
 \toprule
 none& 0.217 $\pm$ 0.004 & 0.177 $\pm$ 0.002 & 0.134 $\pm$ 0.001 & 0.215 $\pm$ 0.002 \\
 \midrule
 0.9 & 0.217 $\pm$ 0.002 & 0.180 $\pm$ 0.002 & 0.134 $\pm$ 0.001 & 0.218 $\pm$ 0.002 \\
 0.8 & 0.220 $\pm$ 0.002 & 0.179 $\pm$ 0.002 & 0.135 $\pm$ 0.003 & 0.220 $\pm$ 0.003 \\
 0.7 & 0.179 $\pm$ 0.097 & 0.148 $\pm$ 0.080 & 0.112 $\pm$ 0.060 & 0.223 $\pm$ 0.003 \\
 0.6 & 0.224 $\pm$ 0.005 & 0.190 $\pm$ 0.005 & 0.143 $\pm$ 0.004 & 0.225 $\pm$ 0.003 \\
 0.5 & 0.232 $\pm$ 0.002 & 0.208 $\pm$ 0.004 & 0.159 $\pm$ 0.003 & 0.239 $\pm$ 0.005 \\
\bottomrule
\end{tabular}
\caption{Multi-task and single-task benchmark $R^2$ results for power conversion efficiency (PCE), band gap (GAP), and highest occupied molecular orbital (HOMO) energy of the \textsc{CEP-2017} benchmark for different ratios of kept nodes in each pooling step (averaged over 5 runs, with 5 random splits). Speedup of pooling runs is measured in terms of elapsed real time compared to the run without pooling.}
\label{cep-table}
\end{table*}

In this section, we propose a regression benchmark for hierarchical models using the 2017 non-fullerene electron-acceptor update~\citep{Lopez_2017} to
the Clean Energy Project molecular library~\citep{Hachmann2011}. 
We refer to this dataset as \textsc{CEP-2017}. 
This dataset was generated by combining molecular fragments from a reference library generating 51256 unique molecules. 
These molecular graphs were then used as input to density functional theory electronic-structure calculations of quantum-mechanical observables (such as GAP and HOMO). 
Restrictions of the crowd-sourced computing platform limited these structures to molecules of 306 electrons or less. 
The direct observables quantities are then used in a physically motivated but empirical Scharber~\citep{Scharber_2006} model to predict power conversion efficiency (PCE). 
This efficiency is the ultimate figure of merit for a new photovoltaic material.

We emphasize that this data, generated with an approximate density functional theory method, and then used in an empirical PCE model, lacks predictive power in terms of design of new materials.  
However a machine learning model built on this data is likely to be transferable to other molecular datasets built on higher level theory (such as coupled-cluster calculations) or experimental ground truth. As we are anticipating this future application of the method, we use the raw (\texttt{\_calc}) values rather than the Gaussian process regressed (to a small experimental dataset) values (\texttt{\_calib}). 

The method of construction of the dataset allows us to highlight the coarse-graining interpretation of the pooling layers introduced in the main text, in terms of the explicit combinatorial building blocks of the non-fullerene electron acceptors.

In Table~\ref{cep-table}, we show multi-task and single-task test set evaluation $R^2$ results for the power conversion efficiency (PCE), the band gap (GAP), and the highest occupied molecular orbital (HOMO) energy. We used a dual-message graph-convolutional model with three graph-convolutional layers with node channel dimensions ${[}512, 512, 512{]}$ and edge channel dimensions ${[}128, 128, 128{]}$ with two interleaved layers of simple pooling. We found our model to be a powerful predictor of both fundamental quantum-mechanical properties (GAP and HOMO), and to a lesser extend the more empirical PCE figure. 
The inclusion of pooling layers resulted in a significant speedup and only a very mild decay in performance.

\subsection{Pooling layers illustrations}
\begin{figure*}[!htp]
\centering
\subfloat{Molecular graphs before pooling}\\{\includegraphics[width=0.8\textwidth]{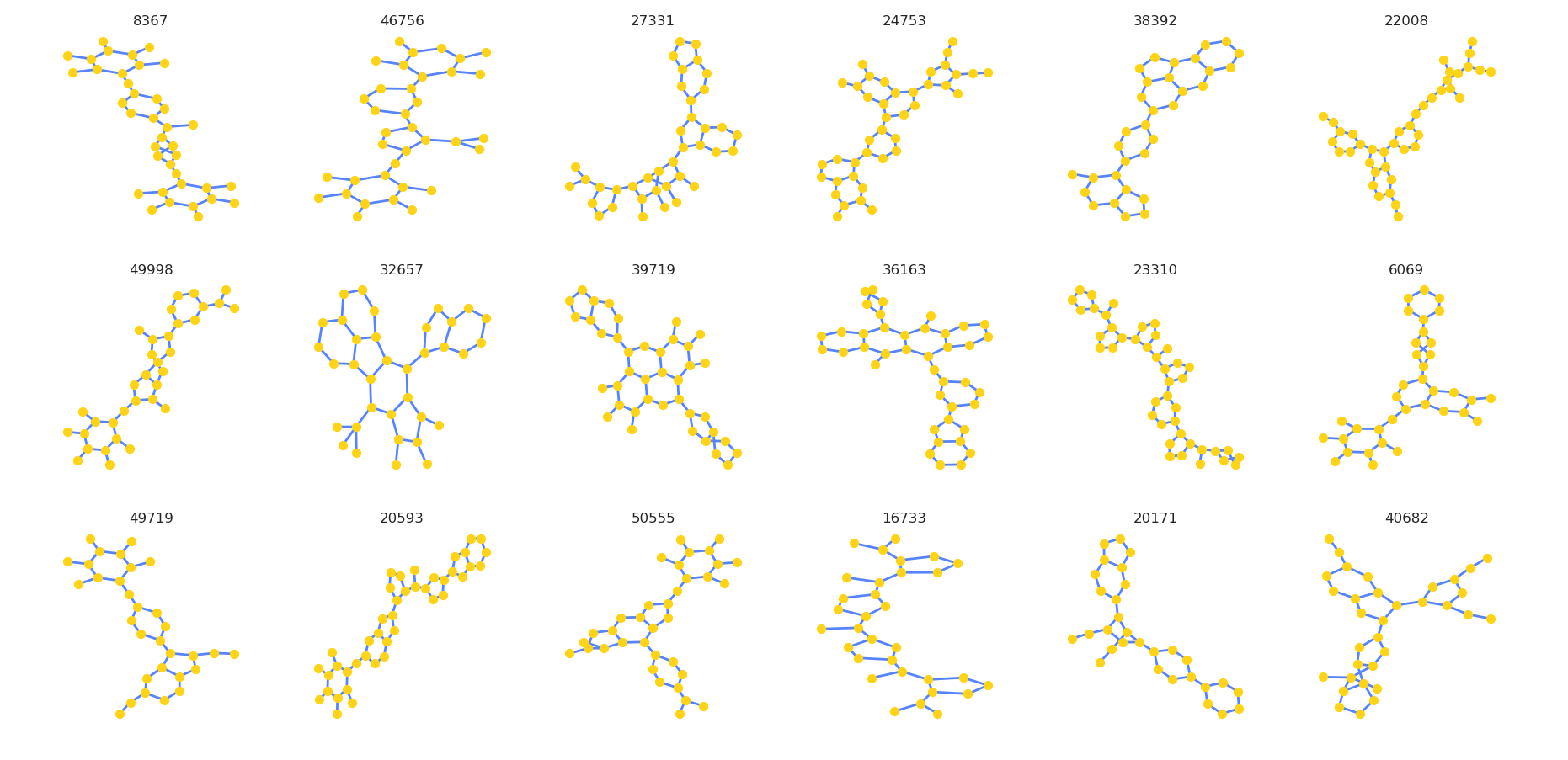}}\\
\subfloat{Coarse-grained graphs after pooling layer 1.}\\{\includegraphics[width=0.8\textwidth]{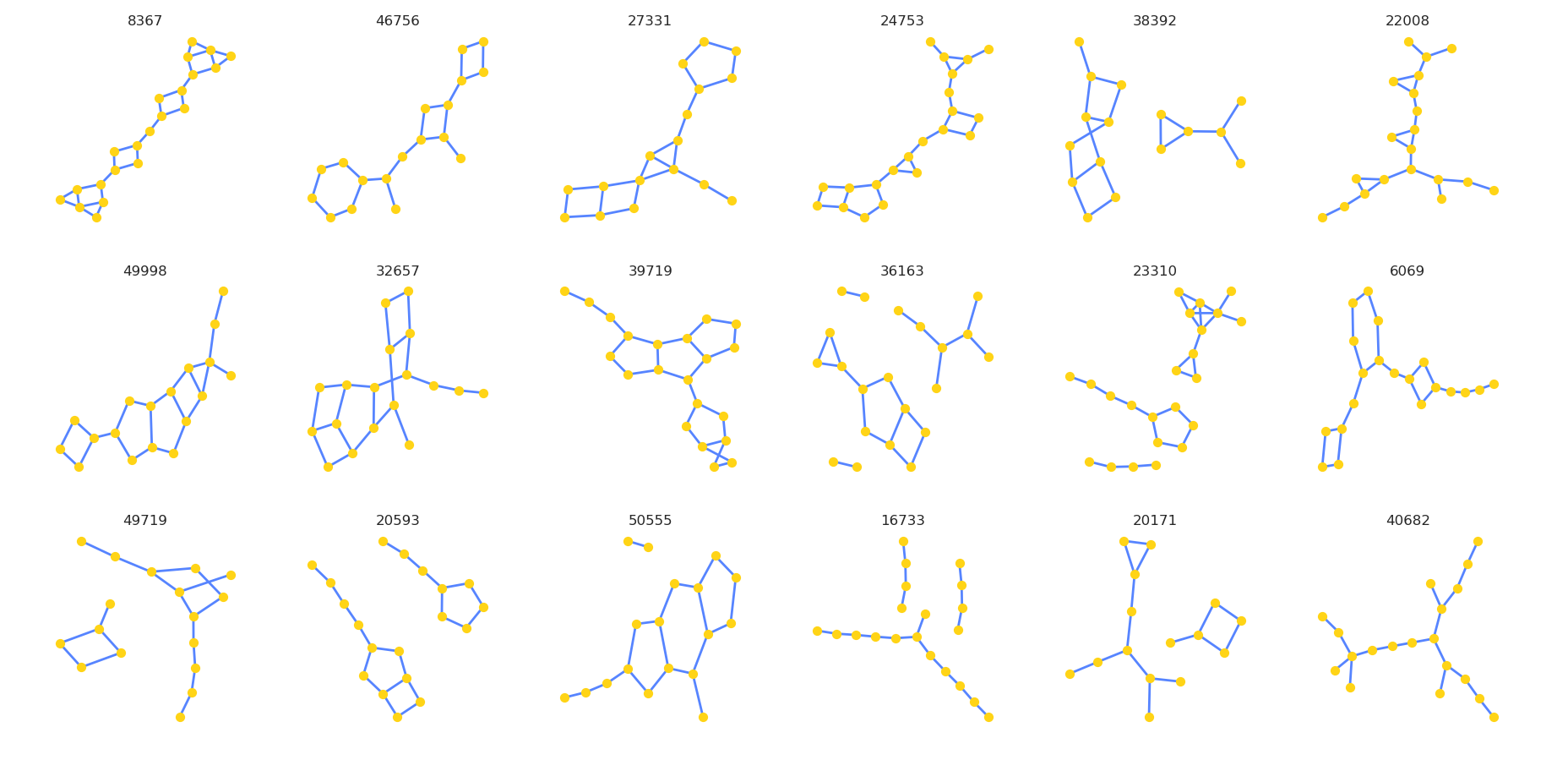}}\\
\subfloat{Coarse-grained graphs after pooling layer 2.}\\{\includegraphics[width=0.8\textwidth]{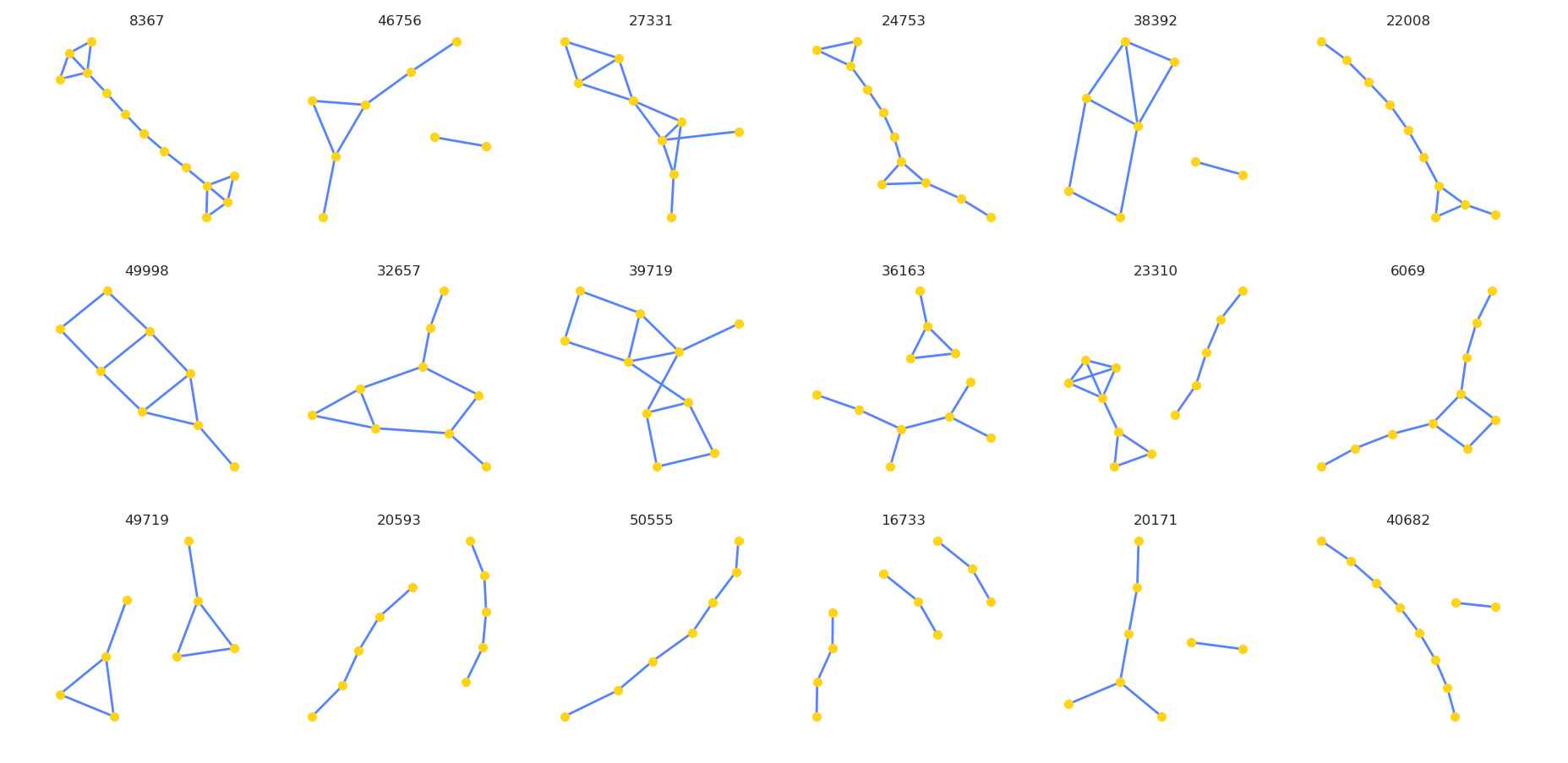}}%
\caption{Pooling of molecular graphs (heavy atoms only) sampled from \textsc{CEP-2017} dataset.}
\label{fig:pooling_examples}
\end{figure*}

In Fig.~\ref{fig:pooling_examples}(a-c) we visualize the effect of two consecutive pooling layers (each keeping only 50\% of the nodes) on a batch of molecules for a \textsc{DM-SimplePooling} model trained on a random split of the \textsc{CEP-2017} dataset introduced in Sec.~\ref{app:cep}. After the first pooling layer (Fig.~\ref{fig:pooling_examples}(b)), the model has approximately learned to group rings and identify the backbones or main connected chains of the molecules. After the second pooling layer (Fig.~\ref{fig:pooling_examples}(c)), the molecular graphs have been reduced to basic, abstract components connected by chains, encoding a coarse-grained representation of the original molecules. Disconnected parts can be interpreted as a consequence of the aggressive pooling forcing the model to pay attention to the parts it considers most relevant for the task at hand.
\newpage
\clearpage

\clearpage
\newpage

\label{app:hyperparams}

\begin{table*}[h!]
\centering
\small
\begin{tabular}{llccc}
 \toprule
 \bf{Dataset} & \bf{Model} & \bf{Keep ratio} & \bf{Node Channels} & \bf{Edge Channels} 
  \\
\toprule
\emph{ESOL} & NoPooling          &     & {[}128, 128{]}      & {[}128, 128{]}      \\
\cmidrule(lr){2-5}
            & CoarseGrainPooling & 0.9 & {[}128, 128{]}      & {[}256, 256{]}      \\
            &                    & 0.8 & {[}128, 128{]}      & {[}64, 64{]}        \\
            &                    & 0.7 & {[}512, 512, 512{]} & {[}128, 128, 128{]} \\
            &                    & 0.6 & {[}256, 256, 256{]} & {[}128, 128, 128{]} \\
            &                    & 0.5 & {[}128, 128{]}      & {[}64, 64{]}        \\
\cmidrule(lr){2-5}
            & SimplePooling      & 0.9 & {[}256, 256{]}      & {[}128, 128{]}      \\
            &                    & 0.8 & {[}256, 256, 256{]} & {[}256, 256, 256{]} \\
            &                    & 0.7 & {[}128, 128, 128{]} & {[}128, 128, 128{]} \\
            &                    & 0.6 & {[}512, 512{]}      & {[}128, 128{]}      \\
            &                    & 0.5 & {[}256, 256{]}      & {[}256, 256{]}      \\
\toprule
\emph{Lipophilicity} & NoPooling          &     & {[}256, 256, 256{]} & {[}64, 64, 64{]}    \\
\cmidrule(lr){2-5}
            & CoarseGrainPooling & 0.9 & {[}256, 256{]}      & {[}128, 128{]}      \\
            &                    & 0.8 & {[}256, 256{]}      & {[}64, 64{]}        \\
            &                    & 0.7 & {[}256, 256{]}      & {[}64, 64{]}        \\
            &                    & 0.6 & {[}256, 256{]}      & {[}256, 256{]}      \\
            &                    & 0.5 & {[}512, 512{]}      & {[}64, 64{]}        \\
\cmidrule(lr){2-5}
            & SimplePooling      & 0.9 & {[}512, 512{]}      & {[}64, 64{]}        \\
            &                    & 0.8 & {[}128, 128{]}      & {[}128, 128{]}      \\
            &                    & 0.7 & {[}256, 256{]}      & {[}128, 128{]}      \\
            &                    & 0.6 & {[}512, 512{]}      & {[}128, 128{]}      \\
            &                    & 0.5 & {[}256, 256{]}      & {[}128, 128{]}      \\
\toprule
\emph{BBBP} & NoPooling          &     & {[}128, 128{]}      & {[}256, 256{]}      \\
\cmidrule(lr){2-5}
            & CoarseGrainPooling & 0.9 & {[}256, 256{]}      & {[}256, 256{]}      \\
            &                    & 0.8 & {[}512, 512{]}      & {[}256, 256{]}      \\
            &                    & 0.7 & {[}128, 128, 128{]} & {[}128, 128, 128{]} \\
            &                    & 0.6 & {[}256, 256{]}      & {[}64, 64{]}        \\
            &                    & 0.5 & {[}256, 256, 256{]} & {[}64, 64, 64{]}    \\
\cmidrule(lr){2-5}
            & SimplePooling      & 0.9 & {[}128, 128, 128{]} & {[}256, 256, 256{]} \\
            &                    & 0.8 & {[}512, 512{]}      & {[}128, 128{]}      \\
            &                    & 0.7 & {[}256, 256{]}      & {[}64, 64{]}        \\
            &                    & 0.6 & {[}512, 512{]}      & {[}256, 256{]}      \\
            &                    & 0.5 & {[}128, 128, 128{]} & {[}64, 64, 64{]}    \\
\toprule
\emph{HIV } & All models         &     & {[}512, 512, 512{]} & {[}128, 128, 128{]} \\
 \bottomrule
\end{tabular}
\caption{Graph-convolutional model hyperparameters used in this work.}
\end{table*}

\end{document}